\documentclass[conference]{IEEEtran}
\IEEEoverridecommandlockouts
\usepackage[utf8]{inputenc}
\usepackage{cite}
\usepackage{amsmath,amssymb,amsfonts}
\usepackage{graphicx, color}
\usepackage{subcaption}
\usepackage[export]{adjustbox}
\usepackage{textcomp}
\usepackage{url} 
\usepackage[table]{xcolor} 
\usepackage{paralist}
\usepackage{enumitem}
\usepackage{multirow}
\usepackage{hyperref}
\usepackage[numbers,sort&compress]{natbib} 
\usepackage{algorithmic}
\usepackage{booktabs}
\usepackage{colortbl}
\usepackage{xcolor} 
\usepackage{booktabs} 
\usepackage{pifont}   
\usepackage{colortbl} 
\usepackage{xcolor}   
\usepackage[linesnumbered,ruled,vlined]{algorithm2e}
\hypersetup{
    colorlinks=true,
    linkcolor=red, 
    citecolor=blue, 
    urlcolor=magenta    
}
\def\ie{{\em i.e.}}
\def\eg{{\em e.g.}}
\def\BibTeX{{\rm B\kern-.05em{\sc i\kern-.025em b}\kern-.08em
    T\kern-.1667em\lower.7ex\hbox{E}\kern-.125emX}}

\newcommand{\smallurl}[1]{\footnotesize\url{#1}}

\definecolor{baselinecolor}{gray}{.9}

\def\sign{\texttt{sign}}

\begin{document}
\title{Robust Diffusion Model-Generated Image Detection with CLIP}
\title{Robust CLIP-Based Detector for Exposing Diffusion Model-Generated Images}



\author{Santosh\textsuperscript{1}, Li Lin\textsuperscript{1}, 
Irene Amerini\textsuperscript{2}, 
Xin Wang\textsuperscript{3}, 
Shu Hu\textsuperscript{1*}\thanks{*Corresponding author}\\
{\textsuperscript{1}Purdue University, {\tt \small \{santosh2, lin1785, hu968\}@purdue.edu} }\\
{\textsuperscript{2}Sapienza University of Rome {\tt \small amerini@diag.uniroma1.it}}\\
{\textsuperscript{3}University at Albany, State University of New York {\tt \small  xwang56@albany.edu}}
}

\maketitle
\thispagestyle{plain}
\pagestyle{plain}

\begin{abstract}
Diffusion models (DMs) have revolutionized image generation, producing high-quality images with applications spanning various fields. However, their ability to create hyper-realistic images poses significant challenges in distinguishing between real and synthetic content, raising concerns about digital authenticity and potential misuse in creating deepfakes. This work introduces a robust detection framework that integrates image and text features extracted by CLIP model with a Multilayer Perceptron (MLP) classifier. We propose a novel loss that can improve the detector's robustness and handle imbalanced datasets. Additionally, we flatten the loss landscape during the model training to improve the detector's generalization capabilities. The effectiveness of our method, which outperforms traditional detection techniques, is demonstrated through extensive experiments, underscoring its potential to set a new state-of-the-art approach in DM-generated image detection. The code is available at \url{https://github.com/Purdue-M2/Robust_DM_Generated_Image_Detection}.


\let\thefootnote\relax\footnote{979-8-3503-7428-5/24/\$31.00 ©2024 IEEE}
\end{abstract}

\section{Introduction}

Diffusion models (DMs) \cite{ho2020denoising, song2020score} are a class of generative models that simulate a stochastic process to transform noise into structured data, effectively producing high-quality images. Unlike traditional generative approaches (\eg, GAN), DMs gradually construct images through an iterative process of denoising, which allows for the generation of images with a high degree of fidelity and detail \cite{dorjsembe2024conditional, hu2024mvdfusion}. This unique method of image creation results in DM-generated images that are not only visually compelling but also versatile in their application across various fields such as health, business, and the creative industries \cite{mei2024segmentation, jodelet2024future}. The advantages of DMs and their generated images lie in their unparalleled ability to produce customized content tailored to specific needs, their flexibility in handling diverse data types, and their effectiveness in tasks that demand high-quality visual representations. This flexibility and high precision offer substantial benefits for enhancing synthetic datasets for machine learning and spearheading innovations in personalized content creation, establishing DMs as a groundbreaking technology in the image generation domain \cite{chen2024deep, jiang2024comat, gal2024lcm}.

\begin{figure}[t]
  \centering
  \includegraphics[width=1\linewidth]{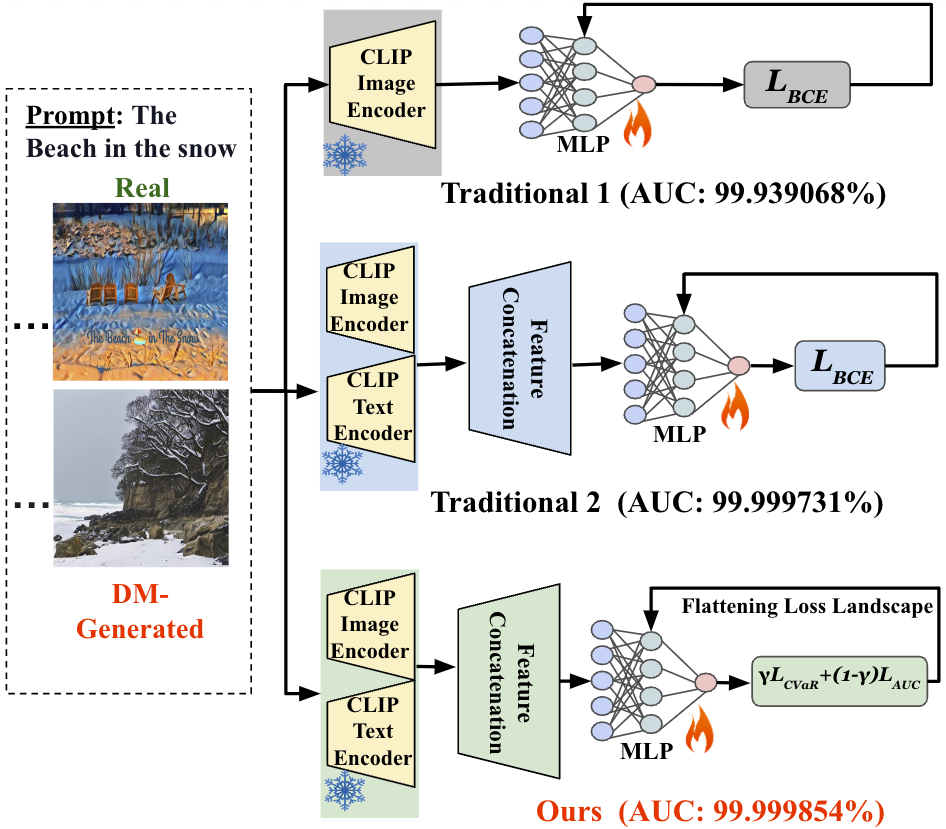}
  \vspace{-5mm}
  \caption{
\textit{Comparison of our method with traditional methods. \textbf{First row:} Traditional Method 1 \cite{cozzolino2023raising} utilizes CLIP image features combined with a Multilayer Perceptron (MLP) classifier and Binary Cross-Entropy (BCE) loss \(L_{\text{BCE}}\). \textbf{Second Row:} Traditional Method 2 \cite{sha2023defake} incorporates both CLIP image and text features with an MLP classifier and BCE loss \(L_{\text{BCE}}\). \textbf{Third row:} Our model enhances DM-generated image detection by using CLIP image and text features, a lightweight MLP classifier, and a combination of Conditional Value at Risk (CVaR) and Area Under the Curve (AUC) losses, across a flattened loss landscape in order to train a robust detector.}}
  \vspace{-5mm}
  \label{fig:introduction}
\end{figure}

Despite the significant advantages of diffusion models (DMs) in generating high-quality, detailed images, their capability poses considerable challenges, particularly in authenticity and trust in digital media \cite{Guarnera2024}. The core disadvantage of DM-generated images lies in their potential misuse for creating highly realistic yet entirely synthetic images, which can be nearly indistinguishable from genuine photographs. This capability raises serious concerns about misinformation, identity theft, and the creation of deepfakes, which can have far-reaching implications in politics, society, and personal security \cite{zhang2021deep, rossler2019faceforensics++}. The convincing nature of these images can be exploited to fabricate evidence, impersonate individuals in sensitive positions, or spread disinformation, undermining trust in digital content and potentially influencing public opinion and personal reputations \cite{marra2018detection}. 

Recent research has emphasized the critical need to tackle this issue, promoting sophisticated analysis methods that capitalize on the unique characteristics of content produced by diffusion models. These include minor irregularities or digital artifacts that do not occur in authentic images \cite{ke2024bilara, luo2024lare2}. These detection capabilities are crucial to mitigate the risks associated with the proliferation of synthetic images, ensuring a more secure and trustworthy digital environment \cite{Khan2024, Lu2024}.
Notably, a range of approaches have been proposed to enhance the detection accuracy and counter the misuse of synthetic imagery. Physical and physiological methods focus on inconsistencies in images, such as aberrant reflections or lighting, and physiological anomalies in human depictions \cite{farid2022lighting, guo2022open, guo2022eyes, hu2021exposing}. Techniques such as analyzing diffuser fingerprints detect unique artifacts inherent to specific image synthesis processes, pointing to their synthetic origins. Additionally, spatial-based and frequency-based methods focus on the textual and spectral anomalies that distinguish fake images from authentic ones \cite{chen2024masked, lin2024detecting, fan2024synthesizing, zhang2023x, yang2023improving, fan2023attacking, wang2022gan}. 

Despite advancements, the robustness of these detection techniques faces significant challenges as generative models improve at evading detection. The reasons can be summarized into four key points: 1) \textbf{Limited Dataset Size:} Detectors are often trained on relatively small datasets, which may not adequately represent the broad diversity of images needed for comprehensive learning. 2) \textbf{Difficulty with Hard Images:} Some images pose greater challenges in learning. Standard training losses, such as cross-entropy, treat all images with equal importance, which can result in inadequate detection of these complex examples.   
3) \textbf{Imbalanced Training Data:} Typically, neural network-based detectors are trained on datasets where DM-generated images outnumber real images, leading to training strategies that might cause the detector to exhibit biased performance by favoring the detection of the majority class. 4) \textbf{Suboptimal Optimization Techniques:} Traditional stochastic gradient descent optimization methods can lead to suboptimal training, where the model settles into local optima, thus reducing its generalization ability across diverse scenarios.
Several recent studies \cite{cozzolino2023raising, sha2023defake} have employed Contrastive Language–Image Pre-training (CLIP) \cite{radford2021learning} ViT as feature extractor, either focusing solely on image representations or integrating both image and text features with Multilayer Perceptrons (MLP) and Binary Cross-Entropy (BCE) loss for the DM-generated images detection. However, these approaches primarily tackle the issue of limited datasets for training. 

To simultaneously address the above four challenges, in this work, we introduce a robust classifier designed to detect DM-generated images. Specifically, we utilize a pre-trained CLIP model to extract both image-level and corresponding text (prompt)-level features for each example. These features are then concatenated to represent the full spectrum of characteristics of the input example. For classification, we employ a trainable MLP model to differentiate between real and DM-generated images. To improve focus on hard examples, we implement a distributionally robust optimization technique known as Conditional Value at Risk (CVaR) loss. To manage the imbalanced training data, we incorporate an Area under the ROC curve (AUC) loss. Additionally, to boost the model's generalization capability, we use Sharpness-Aware Minimization (SAM) to optimize model parameters. A comparative overview of our method against traditional approaches is illustrated in Figure \ref{fig:introduction}.
The main contributions of this work can be summarized as follows:
\begin{itemize}
    \item We propose a lightweight, robust detector to detect images generated by diffusion models effectively. 
    \item Our method enhances robustness comprehensively by establishing a framework that operates across three critical levels: features, loss, and optimization.  
    \item Our method outperforms the state-of-the-art approaches, as highlighted in the extensive experiments and results.
\end{itemize}

\section{Related Work}
\subsection{DM-generated Image}
Advancements in diffusion models (DMs) have led to their application in a wide array of generative tasks, demonstrating unparalleled ability in synthesizing realistic images and content. Following foundational work, Stable Diffusion (SD) \cite{Rombach_2022_CVPR} emerged as a versatile framework, enabling detailed image generation from textual prompts with various iterations including SD v1.4 and SD v2.0. Latent Diffusion Models (LDMs) represent a significant leap, reducing computational demands by operating in a latent space, thus facilitating higher resolution outputs. Additionally, commercial platforms have begun integrating these technologies, with DALL·E 2 by OpenAI \cite{ramesh2022dalle2} and Imagen by Google \cite{saharia2022imagen} leading the charge in text-to-image generation, pushing the boundaries of creative potential and application. Models such as Guided Diffusion~\cite{dhariwal2021guided} and Cascaded Diffusion \cite{nichol2021improved} employ nuanced approaches to refine the synthesis process, thereby reducing artifacts and enhancing the naturalness of generated images. The introduction of Midjourney~\cite{midjourney} and the expansion of Stable Diffusion into user-friendly interfaces like DreamStudio highlight the transition of DMs from research to mainstream use, indicating a growing interest and investment in generative AI technologies. These models, from their academic roots to commercial implementations, illustrate the dynamic evolution of diffusion models, setting new standards for artificial intelligence in creative domains.

\begin{figure*}[t]
    \centering
    \includegraphics[width=1\textwidth]{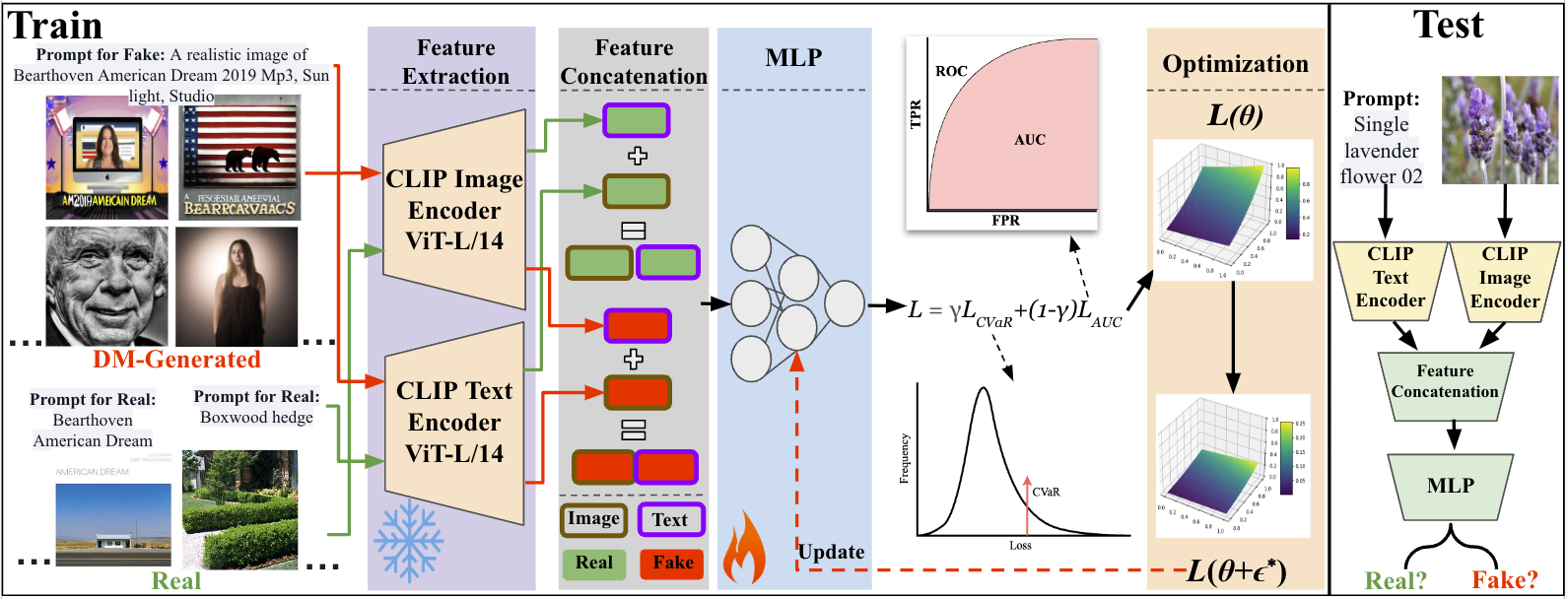}
    \caption{\textit{Overview of our proposed model using CLIP for encoding the input images and text, concatenating the image and text features, an MLP module with robust CVaR + AUC loss, and an optimization step involving a flattened loss landscape for detecting DM-generated images apart from the real images. The snowflake represents the module is fixed. The fire means the module will be trained.}}
    \label{fig:galaxy}
\end{figure*}
\subsection{DM-generated Image Detection}
In the field of synthetic image detection, especially for images generated by diffusion models (DMs), methodologies can be distinctly categorized into two primary approaches: methods from the first category analyze solely the image data, and methods from the second category incorporate both images and textual information for classifying the input images.
Specifically, the first category of methods primarily focuses on analyzing image data, employing a range of sophisticated techniques specifically tailored for diffusion models (DMs), and in some instances, methodologies originally developed for generative adversarial networks (GANs) have been effectively adapted to address the distinct challenges posed by DM-generated images. For example,  \cite{borji2023qualitative} focuses on physical/physiological features for classifying the real vs. fake content by categorizing qualitative failures of image generation models into five main types (\eg, human parts, geometry, physics) and identifying common failure points such as unnatural textures or incorrect geometries. Similarly, \cite{farid2022perspective} assesses if the 3D structures, shadows, and reflections in generated images conform to the physical and geometric rules that typically govern real-world photography and \cite{farid2022lighting} applies a spherical harmonic-based approach to analyze the lighting in images generated by DALL-E-2 from text prompts. The specified focus on the datasets in these methodologies limit their generalization capabilities. 
\cite{sinitsa2023deep} introduces a low-resource method for synthetic image detection and model lineage analysis that uses deep image fingerprints derived from convolutional neural network (CNN) architectures. Similarly, \cite{ma2023exposing} integrates stepwise error analysis at multiple stages of the diffusion process. Deep neural network-based and transformer-based classifiers \cite{huang2017, szegedy2016, chollet2017, cozzolino2023raising} have proven effective in detecting images manipulated by image-to-image translation networks. However, their application to DM-generated images is often limited by challenges in generalizing to content developed by unseen models. 

In the second category, methods utilize both visual and textual data, thereby broadening the scope of analysis. This dual-input strategy significantly enhances the detection capabilities by combining the strengths of image and text-based features.
A notable implementation of this approach is \cite{sha2023defake}, which employs CLIP’s dual encoders to process images alongside their corresponding textual prompts. The visual and textual features are then concatenated to form a composite input.  This method exemplifies the enhanced sensitivity and accuracy that can be achieved in classifying DM-generated content by addressing its inherent multimodal nature. However, this method does not consider robustness problems in real-world scenarios, which is the focus of our proposed work.

\section{Method}
Figure \ref{fig:galaxy} provides an overview of our proposed framework.

\subsection{Feature Space Modeling}
We design our classification algorithm based on the image and text features encoded by the CLIP (Contrastive Language-Image Pre-training)  ViT L/14 model, which has been trained on a large dataset containing 400 million image-text pairs \cite{radford2021learning}. 
The significance of this module lies in its ability to enable the model to identify both high-level semantic and detailed visual features, which are crucial for the classification task without requiring a large dataset for training. Consequently, we can ensure that our method is both lightweight and highly effective, facilitating easier deployment.

For a Dataset $D = {(X_i, T_i, Y_i)}_{i=1}^{n}$, each element comprises an image $X_i$, its associated text $T_i$, and a binary label $Y_i \in \{0,1\}$, where 0 represents an authentic image and 1 a diffusion-generated image. By processing each image-text pair through the CLIP model, we obtain a joint feature representation, which incorporates the attributes essential for distinguishing between the two types of images. This approach generates a 1536-dimensional feature vector for each dataset entry, equally divided between image-derived and text-derived features. These vectors are collected into a feature bank $C = {(F_i, Y_i)}_{i=1}^{n}$, where $F_i$ is the combined feature vector of the $i$-th dataset entry, and $Y_i$ is its label. 

\subsection{MLP Classifier Architecture}
Upon establishing the feature bank, we introduce a 3-layer Multilayer Perceptron (MLP) as our classifier. To ensure stability and generalization, batch normalization and ReLU activation follow each linear layer, with dropout included to prevent overfitting. The classifier's output layer adapts to our binary classification task using a single neuron with sigmoid activation.

\subsection{Objective Function}

To effectively differentiate real images from diffusion-generated ones, we propose a dual-objective loss framework that synergizes Conditional Value-at-Risk (CVaR) and an AUC loss. This framework is designed to tackle the challenges posed by hard examples and imbalanced training sets.

\textbf{Conditional Value-at-Risk (CVaR) Loss.} The CVaR loss \cite{lin2024robust, lin2024robust2, lin2024preserving, ju2024improving, hu2024outlier, hu2023rank, hu2022distributionally, hu2021tkml, hu2022sum, hu2020learning} component is formulated to make the model focus on the hardest examples in the dataset, thereby improving robustness. It is defined as:

\begin{equation}
L_{CVaR}(\theta) = \min_{\lambda \in \mathbb{R}} \left\{ \lambda + \frac{1}{\alpha n} \sum_{i=1}^{n} \left[ \ell(\theta; F_i, Y_i) - \lambda \right]_+ \right\},
\label{eq:cvar}
\end{equation}
where $\ell$ is the classification loss function (\eg, binary cross-entropy loss), $\theta$ symbolizes the model parameters, and $(F_i,Y_i)$ are the feature-label pairs from the dataset. $[a]_+ = \max\{0, a\}$ is the hinge function. The parameter $\alpha \in (0,1)$ facilitates tuning the focus from the most challenging to more average examples. 

\textbf{AUC loss.} For enhancing the model's discrimination ability, especially under class imbalance, an area under the curve (AUC) loss \cite{pu2022learning, guo2022robust} component is utilized. This component is strategically developed to optimize the AUC metric directly, ensuring the model effectively ranks pairs of positive and negative examples. Suppose $P=\{i|y_i=+1\}$ and $N=\{i|y_i=0\}$ are sets of indices for positive (diffusion-generated) and negative (real) examples, respectively. Then, the AUC loss can be written as follows:

\begin{equation}
L_{AUC}(\theta) = \frac{1}{|P||N|} \sum_{i \in P} \sum_{j \in N} R(s(\theta; F_i), s(\theta; F_j)),
\end{equation}
where
\begin{equation}\small
\begin{aligned}
&R(s(\theta; F_i), s(\theta; F_j)) \\
&=\left\{\begin{matrix}
(-(s(\theta; F_i)-s(\theta; F_j) - \eta))^p, & s(\theta; F_i)-s(\theta; F_j)< \eta ,\\ 
 0,& \mbox{otherwise},
\end{matrix}\right.
\end{aligned}
\label{eq:gamma}
\end{equation}
$\eta \in (0,1]$ and $p > 1$ are two hyperparameters. $s(\theta; F)$ indicates the score function of the model (\ie, MLP) parameterized by $\theta$ for feature $F$. This formulation is aimed at increasing the margin between scores of positive and negative examples, thereby enhancing the AUC metric's contribution directly.

\textbf{Total Loss Function.} The overarching training objective harmonizes the CVaR and AUC losses into a unified framework:
\vspace{-1mm}
\begin{equation}
\vspace{-1mm}
L(\theta) = \gamma L_{CVaR}(\theta) + (1 - \gamma) L_{AUC}(\theta),
\label{eq:total_loss}
\end{equation}
where $\gamma$ serves as a balancing parameter to modulate the contributions from both loss components. This compound loss function enables targeted learning from the most challenging instances, as prioritized by the CVaR component, while also promoting an improved classification performance that is not affected by the imbalanced data, as facilitated by the AUC-related component. The integrated approach provides a comprehensive solution to model optimization, tailored to the nuanced requirements of differentiating real from diffusion-generated images.

\subsection{Optimization Method}
Enhancing the model's generalization capabilities is achieved through Sharpness-Aware Minimization (SAM) \cite{foret2020sharpness}, which seeks a flat loss landscape conducive to robust classification across diverse scenarios. This is realized by optimizing for a perturbation $\epsilon^*$ that maximizes the  loss $L(\theta)$ within a defined perturbation magnitude $\delta$:

\begin{equation}
\begin{aligned}
\epsilon^* &= \arg\max_{\|\epsilon\|_2 \leq \delta} L(\theta)\approx\arg\max_{\|\epsilon\|_2\leq \delta}\epsilon^\top\nabla_\theta L=\delta\sign(\nabla_\theta L),
\end{aligned}
\label{eq:epsilon}
\end{equation}

The approximation is obtained using first-order Taylor expansion with the assumption that $\epsilon$ is small. The final equation is obtained by solving a dual norm problem, where $\sign$ represents a sign function and $\nabla_\theta L$ is the gradient of $L$ with respect to $\theta$. As a result, the model parameters are updated by solving the following problem:
\vspace{-1mm}
\begin{equation}
    \begin{aligned}
        \min_{\theta} L\textbf{(}\theta+\epsilon^*\textbf{)}.
    \end{aligned}
\label{eq:sharpness}
\vspace{-3mm}
\end{equation}
\vspace{-2mm}
\begin{algorithm}[t!]
    \caption{End-to-end Training}\label{alg:Optimization}
    \begin{algorithmic}[1]
        \REQUIRE A training dataset $C$ with size $n$, $\gamma$,  $\alpha$, $\eta$, $p$, max\_iterations, num\_batch, learning rate $\beta$
        \ENSURE Optimized parameters
        
        \STATE \textbf{Initialization:} $\theta_0$, $l=0$
        
        \FOR{$e=1$ to \textit{max\_iterations}}
            \FOR{$b=1$ to \textit{num\_batch}}
                \STATE Sample a mini-batch $C_b$ from $C$
                \STATE Compute $\ell(\theta_l;F_i,Y_i)$, $\forall (F_i,Y_i)\in C_b$
                \STATE Use binary search to find $\lambda$ that minimizes $L$ (Eq. \ref{eq:total_loss}) on $C_b$
                \STATE Compute $\epsilon^*$ based on Eq.(\ref{eq:epsilon})
                \STATE Compute gradient approximation for the total loss
                \STATE Update $\theta$: $\theta_{l+1} \leftarrow \theta_l - \beta \nabla_\theta L \big|_{\theta_l+\epsilon^*}$
                \STATE $l \leftarrow l+1$
            \ENDFOR
        \ENDFOR
        \STATE \textbf{return} $\theta_l$ \COMMENT{Return the optimized model parameters}
    \end{algorithmic}
\end{algorithm}


\textbf{End-to-end Training}. The training begins with the initialization of the model parameters $\theta$. For each training iteration, a mini-batch $C_b$ is randomly selected from the comprehensive feature bank $C$. The steps involved in each iteration are laid out as follows (also see Algorithm \ref{alg:Optimization}):

\begin{compactitem}
\item Fix $\theta$ and use binary search to find the global optimum of $\lambda$ since (\ref{eq:total_loss}) is convex with respect to $\lambda$.
\item Fix $\lambda$, compute $\epsilon^*$ based on Eq. (\ref{eq:epsilon}).
\item Update $\theta$ based on the gradient approximation for (\ref{eq:sharpness}):  $\theta \leftarrow \theta - \beta \nabla_{\theta} L(\theta + \epsilon^*)$, where $\beta$ is the learning rate.
\end{compactitem}

\section{Experiments}

\subsection{Experimental Settings}

\textbf{Datasets.} Our experiments are based on the Diffusion-generated Deepfake Detection dataset (D3) \cite{D3} which includes a comprehensive collection of approximately 2.3 million records and 11.5 million images in the training set. This dataset comprises real images sourced from the LAION-400M \cite{schuhmann2021laion400m} dataset and four synthetic images for each entry. The synthetic images are produced from a single textual prompt using a diverse array of text-to-image generation models, namely Stable Diffusion 1.4 (SD-1.4) \cite{Rombach_2022_CVPR}, Stable Diffusion 2.1 (SD-2.1) \cite{Rombach_2022_CVPR}, Stable Diffusion XL (SD-XL) \cite{stableDiffXL2023}, and DeepFloyd IF (DF-IF) \cite{deepfloyd2023}. This selection provides a broad representation of current synthetic image generation capabilities, including both variations of the Stable Diffusion technique and the distinct approach inspired by Imagen, represented by DeepFloyd IF. The dataset showcases a variety of aspect ratios, specifically $256 \times 256$, $512 \times 512$, $640 \times 480$, and $640 \times 360$. However, we resize all images to $224 \times 224$. It also utilizes a broad spectrum of encoding and compression formats to mirror the prevalent image distribution found, with a focus on JPEG encoding to match the common distribution within the LAION dataset. The test set follows the similar format and includes a smaller set of 4,800 records, consisting of 4,800 real images and 19,200 synthetic images from the aforementioned generative models.

\textbf{Evaluation Metrics.} The model's effectiveness is measured by the Area Under the ROC Curve (AUC). This metric is utilized to measure the model's discriminative performance, serving as the primary benchmark for comparison.

\textbf{Baseline Methods.}
We benchmark our approaches against two baseline models that utilize the CLIP architecture for feature extraction in the detection of AI-generated images. 
1) The first baseline \cite{cozzolino2023raising} employs a combination of CLIP image features with a Multilayer Perceptron (MLP) and Binary Cross-Entropy (BCE) loss for binary classification of images. 
2) The second baseline \cite{sha2023defake} extends this approach by integrating both the visual and textual features provided by CLIP. It processes images alongside their corresponding textual prompts through CLIP’s dual encoders, concatenating these features to form a composite input for the MLP classifier, also utilizing BCE loss. We implement and conduct extensive testing on both the baseline models for a thorough assessment of their performance in the same dataset.

\textbf{Implementation Details.}
Our method and the second baseline employ a hybrid approach, integrating visual and textual features within a singular model framework. This integration results in a feature space of 1536 dimensions (768 for visual features, 768 for textual features) per data instance. All model architectures comprise three Multi-Layer Perceptron (MLP) layers, each equipped with 1536 units to navigate the complexities of the enhanced feature space.
We conduct the training on a batch size of 32, optimizing data throughput for each iteration. An Adam optimizer, with an initial learning rate of $\beta=1e-3$, is utilized, complemented by a Cosine Annealing Learning Rate Scheduler to adaptively refine the learning rate throughout the training process. $\delta$ is 0.05 as specified in (\ref{eq:epsilon}). 
Tuning the hyperparameters, we experiment the $\alpha$ values for CVaR Loss from 0.1 to 0.9. Once the best $\alpha$ hyperparameter is found, we experiment with trade-off hyperparameter $\gamma$ from 0.1 to 0.9 with $\eta$ and $p$ set to 0.6 and 2,  respectively. The implementation leverages the PyTorch framework, with computational support from an NVIDIA RTX A6000 GPU. 

\subsection{Results}
We report the performance of our method and the compared methods in Table \ref{tab:model_performance}. It is clear that our method outperforms the baselines, which achieves an outstanding AUC of 99.999854\%, demonstrating near-perfect discrimination between real and DM-generated images. 

\begin{table}[t!]
\centering
\caption{Comparison with the baseline methods in terms of AUC score. The best results are shown in \textbf{Bold}.}
\label{tab:model_performance}
\begin{tabular}{@{}c|c|c|c@{}}
\toprule
Metric & Traditional 1 \cite{cozzolino2023raising} &  Traditional 2 \cite{sha2023defake} & Ours\\ \midrule
AUC(\%) & 99.939068 & 99.999731 & \textbf{99.999854} \\
\bottomrule
\end{tabular}
\end{table}

\subsection{Ablation Study}
\textbf{Effects of CVaR, AUC, and SAM.} In this ablation study, we systematically examine and compare the performance of variants (V1, V2, V3, and V4) of our method against our final method to demonstrate the effectiveness of each module. As detailed in Table \ref{tab: Ablation Study}, the performance comparison between V4 and ours shows an improvement of 0.0005\%, indicating the effectiveness of AUC loss. Comparing V3 with ours, the AUC increases by 0.0107\%, highlighting the impact of SAM. Additionally, the transition from V2 to V3, which incorporates the CVaR component, results in a 0.028\% improvement in AUC. Therefore, the impact of each component in our method is: CVaR loss $>$ SAM $>$ AUC loss. 

\newcommand{\cmark}{\ding{51}} 
\begin{table}[t!]
\centering
\caption{An ablation study of key components in our framework.}
\label{tab: Ablation Study}
\begin{tabular}{@{}c|c|c|c|c@{}}
\toprule
Method & CVaR Loss & AUC Loss & SAM & AUC (\%) \\
\midrule
V1     & \cmark    &          &     & 99.999764 \\
\midrule
V2     &           & \cmark   &     & 99.999467 \\
\midrule
V3     & \cmark    & \cmark   &     & 99.999747 \\
\midrule
V4     & \cmark    &          &\cmark&99.999849 \\
\midrule
Ours   & \cmark    & \cmark   & \cmark & \textbf{99.999854} \\ \hline
\end{tabular}
\vspace{-2mm}
\end{table}

\begin{figure}[ht]
    \centering
    \includegraphics[width=\linewidth]{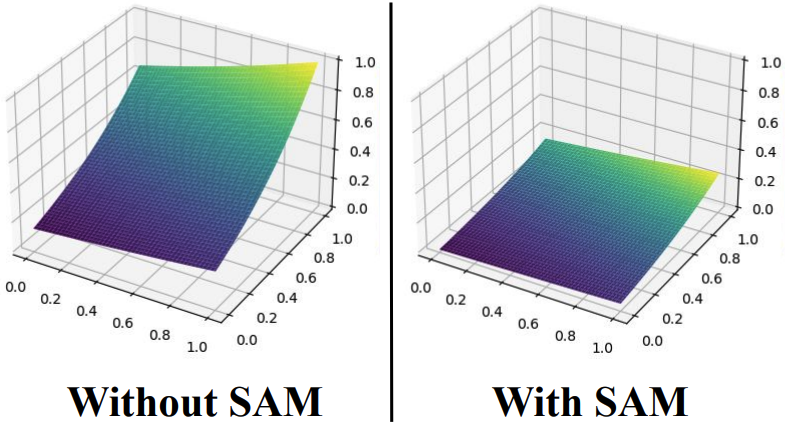}
    \caption{The loss landscape visualization of our proposed method without (left) and with (right) using the Sharpness Aware minimization (SAM) method.}
    \label{fig:loss_landscapes}
    \vspace{-2mm}
\end{figure}

\textbf{Visualization of Loss Landscape.} 
Figure \ref{fig:loss_landscapes} clearly demonstrates the benefits of incorporating SAM optimization into our methodology. The absence of SAM results in a complex and uneven loss landscape, complicating the optimization efforts and potentially causing erratic generalization. On the other hand, the figure on the right illustrates a significantly smoother loss landscape when SAM is integrated. This smoother landscape implies a more stable model that is better at generalizing to unfamiliar data. 

\subsection{Sensitivity Analysis}

\begin{figure}[t!]
    \centering
    \includegraphics[width=\linewidth]{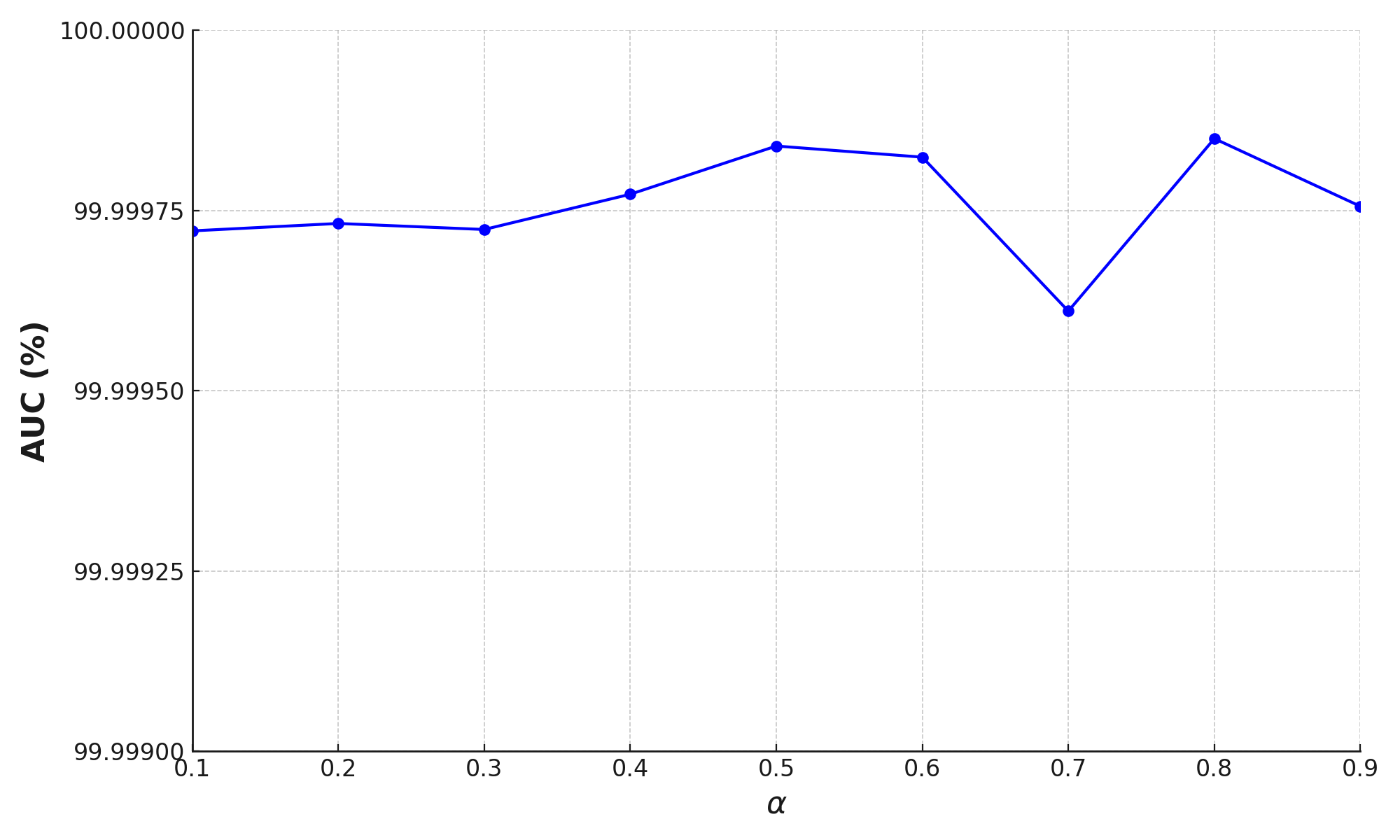}
    \caption{AUC score with respect to different $\alpha$ values.}
    \label{fig:CVaR_Performance}
    \vspace{-4mm}
\end{figure}

In the sensitivity analysis, the influence of $\alpha$ in the CVaR loss is examined. The range of $\alpha$ values from 0.1 to 0.9 was explored to identify the optimal setting for CVaR, crucial for maximizing detection capabilities. Figure \ref{fig:CVaR_Performance} illustrates the AUC as a function of various CVaR $\alpha$ values. This analysis pinpointed an optimal CVaR $\alpha$ value of 0.8, correlating with the highest AUC. 

\begin{figure}[t!]
    \centering
    \includegraphics[width=\linewidth]{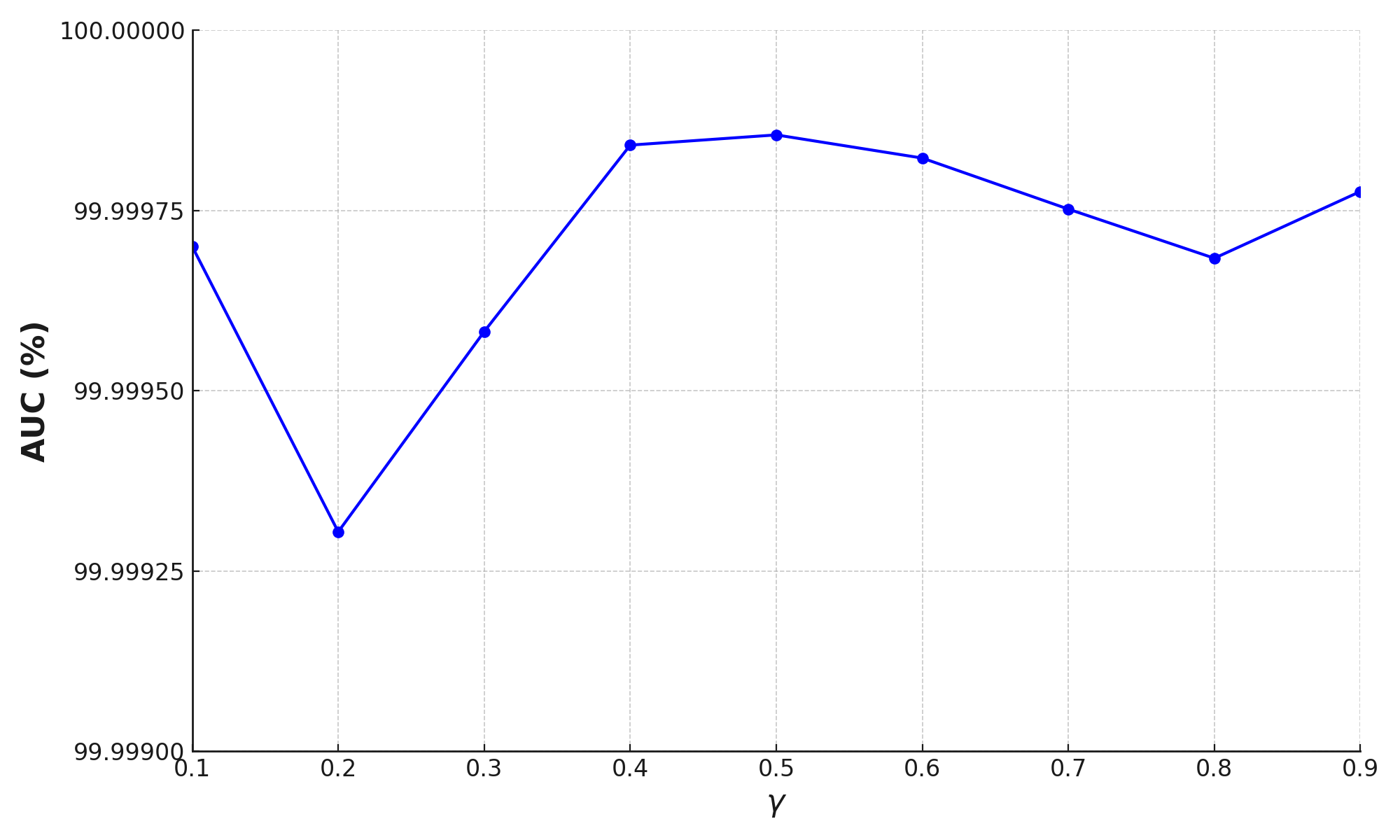}
    \caption{AUC score with respect to different $\gamma$ values by fixing $\alpha=0.8$.}
    \label{fig:AUC_performance}
    \vspace{-5mm}
\end{figure}

Following the determination of the optimal CVaR $\alpha$ at 0.8, a further analysis was conducted to evaluate the trade-off between CVaR loss and AUC loss in Eq. (\ref{eq:total_loss}). The range of $\gamma$ values from 0.1 to 0.9 was revisited for this purpose. The results, presented in Figure \ref{fig:AUC_performance}, indicated that an $\gamma$ of 0.5 for the CVaR and AUC trade-off facilitated the most substantial improvement in model performance.

\section{Conclusion}

The challenge of accurately detecting images generated by diffusion models (DMs) is compounded by the sophistication of the generative processes and the realism of the output images. Addressing these issues, we introduce a robust detection framework that employs CLIP image and text features combined with a Multilayer Perceptron (MLP) classifier. We utilize a combination of CVaR and AUC losses for model training. The model is then trained with a loss landscape flattening strategy. Our approach not only enhances the model's robustness but also improves the model's generalization capability, which has demonstrated superior capabilities in distinguishing between real and DM-generated images, as confirmed by extensive experiments that yielded nearly perfect discrimination capabilities.

\textbf{Limitations.} One limitation is that our method can only test images that have corresponding text information.  

\textbf{Future Work.} We plan to broaden the scope of our methodology by incorporating additional modalities and datasets to demonstrate the robustness and scalability of our detection techniques. We also aim to extend our approach to address newer generative models beyond DMs, such as advanced versions of GANs.

\smallskip
\smallskip
\noindent\textbf{Acknowledgments.} This work is supported by the U.S. National Science Foundation (NSF) under grant IIS-2434967 and the National Artificial Intelligence Research Resource (NAIRR) Pilot and TACC Lonestar6.
The views, opinions and/or findings expressed are those of the author and should not be interpreted as representing the official views or policies of NSF and NAIRR Pilot.
{\small
\bibliographystyle{unsrt}
\bibliography{main}
\end{document}
}